\documentclass[sigconf]{acmart}

\AtBeginDocument{%
  }

\setcopyright{acmlicensed}
\copyrightyear{2018}
\acmYear{2018}
\acmDOI{XXXXXXX.XXXXXXX}
\acmConference[Conference acronym 'XX]{Make sure to enter the correct
  conference title from your rights confirmation email}{June 03--05,
  2018}{Woodstock, NY}
\acmISBN{978-1-4503-XXXX-X/2018/06}

\makeatletter
\def\@ACM@checkaffil{} 
\makeatother

\usepackage[utf8]{inputenc} 
\usepackage[T1]{fontenc}    
\usepackage{hyperref} 
\usepackage{url}            
\usepackage{booktabs}       
\usepackage{amsfonts}       
\usepackage{nicefrac}       
\usepackage{microtype}      
\usepackage{xcolor}         
\usepackage{multirow}
\usepackage{fontawesome}
\usepackage{graphicx}
\usepackage{xspace}
\usepackage{graphicx}
\usepackage{amsmath}
\usepackage{amssymb}
\usepackage{booktabs}
\usepackage{algorithm}
\usepackage[table]{xcolor}
\usepackage{algorithmic}
\usepackage{multirow}
\usepackage{siunitx}
\copyrightyear{2025}
\acmYear{2025}
\setcopyright{acmlicensed}
\acmConference[MM '25]{Proceedings of the 33rd ACM International Conference on Multimedia}{October 27--31, 2025}{Dublin, Ireland}
\acmBooktitle{Proceedings of the 33rd ACM International Conference on Multimedia (MM '25), October 27--31, 2025, Dublin, Ireland}
\acmDOI{10.1145/3746027.3755141}
\acmISBN{979-8-4007-2035-2/2025/10}

\begin{document}

\title{Twin Co-Adaptive Dialogue for Progressive Image Generation}

\author{Jianhui Wang}
\authornote{Equal contribution.}
\affiliation{%
  \institution{University of Electronic Science and Technology of China}
}

\author{Yangfan He}
\authornotemark[1]
\affiliation{%
  \institution{University of Minnesota Twin Cities}
}

\author{Yan Zhong}
\affiliation{%
  \institution{Peking University}
}

\author{Xinyuan Song}
\affiliation{%
  \institution{Emory University}
}

\author{Jiayi Su}
\affiliation{%
  \institution{Xiamen University Malaysia}
}

\author{Yuheng Feng}
\authornote{Project lead.}
\affiliation{%
  \institution{Hong Kong Polytechnic University}
}

\author{Ruoyu Wang}
\affiliation{%
  \institution{Tsinghua University}
}

\author{Hongyang He}
\affiliation{%
  \institution{University of Warwick}
}

\author{Wenyu Zhu}
\affiliation{%
  \institution{Tsinghua University}
}

\author{Xinhang Yuan}
\affiliation{%
  \institution{Washington University, Saint Louis}
}

\author{Miao Zhang}
\authornote{Corresponding author.}
\affiliation{%
  \institution{Tsinghua University}
}

\author{Keqin Li}
\affiliation{%
  \institution{University of Toronto}
}

\author{Jiaqi Chen}
\affiliation{%
  \institution{Google}
}

\author{Tianyu Shi}
\authornotemark[3]
\affiliation{%
  \institution{University of Toronto}
}

\author{Xueqian Wang}
\affiliation{%
  \institution{Tsinghua University}
}

\begin{abstract}
Modern text-to-image generation systems have enabled the creation of remarkably realistic and high-quality visuals, yet they often falter when handling the inherent ambiguities in user prompts. In this work, we present Twin-Co, a framework that leverages synchronized, co-adaptive dialogue to progressively refine image generation. Instead of a static generation process, Twin-Co employs a dynamic, iterative workflow where an intelligent dialogue agent continuously interacts with the user. Initially, a base image is generated from the user's prompt. Then, through a series of synchronized dialogue exchanges, the system adapts and optimizes the image according to evolving user feedback. The co-adaptive process allows the system to progressively narrow down ambiguities and better align with user intent. Experiments demonstrate that Twin-Co not only enhances user experience by reducing trial-and-error iterations but also improves the quality of the generated images, streamlining creative process across various applications.
\end{abstract}

%
%




\maketitle

\section{Introduction}

Generative artificial intelligence has rapidly become a pivotal force in driving economic advancement by enhancing a broad range of both creative and routine tasks. Cutting-edge models such as DALL·E 3~\cite{betker2023improving}, Imagen~\cite{saharia2022photorealistictexttoimagediffusionmodels}, Stable Diffusion~\cite{esser2024scalingrectifiedflowtransformers}, and Cogview~3~\cite{zheng2024cogview3finerfastertexttoimage} have substantially raised the bar in producing diverse, photorealistic, and high-fidelity images from textual inputs~\cite{gozalo2023chatgpt}. Yet, despite these significant achievements, critical challenges persist. Existing systems frequently falter in capturing the nuanced intent behind user prompts, resulting in outputs that do not fully align with user expectations. The ongoing difficulty of interpreting subtle user intentions and incorporating iterative feedback highlights the need for innovative, dialogue-driven approaches in image generation.

At the user level, the task of formulating effective prompts presents its own set of challenges. Non-expert users frequently lack the specialized knowledge required to fine-tune input variables, which can result in outputs that diverge significantly from their envisioned goals. Moreover, even when the same prompt is used, variations in content, layout, background, and color often occur, forcing users into a laborious, trial-and-error process. This unpredictability not only amplifies the effort needed to achieve a satisfactory result but also emphasizes the necessity for a more intuitive, interactive system capable of bridging the gap between user intent and the model’s rendering capabilities.

To address these challenges, we propose Twin-Co, a twin co-adaptive dialogue system for progressive image generation. Twin-Co leverages two complementary feedback pathways. One pathway actively engages users through multi-turn, interactive dialogue, capturing their nuanced intents. The other pathway harnesses an internal refinement process that continuously optimizes the generated images. By integrating these twin feedback loops, Twin-Co iteratively aligns image outputs with user intentions, thereby reducing reliance on laborious trial-and-error iterations while streamlining the creative process and enhancing both quality and relevance.

We validate the effectiveness of Twin-Co through extensive experiments across a range of image generation tasks. Our experimental results demonstrate that Twin-Co effectively improves user experience. The framework shows considerable promise in transforming creative workflows by bridging the gap between raw user intent and the final visual output.

Our main contributions are summarized as follows:
\begin{itemize}
    \item We develop novel human-machine interaction techniques tailored for interactive image generation, guiding non-expert users through a refined process that accurately captures and translates their intentions into visual outputs.
    \item We introduce Twin-Co, a twin co-adaptive dialogue framework that integrates multi-turn user feedback with an internal optimization process, leading to progressive image enhancement.
    \item We demonstrate the versatility of Twin-Co across diverse image generation scenarios, underscoring its potential to revolutionize creative workflows through rapid visualization and iterative refinement.
\end{itemize}

\begin{figure}[!t]
    \centering
    \includegraphics[width=0.9\linewidth]{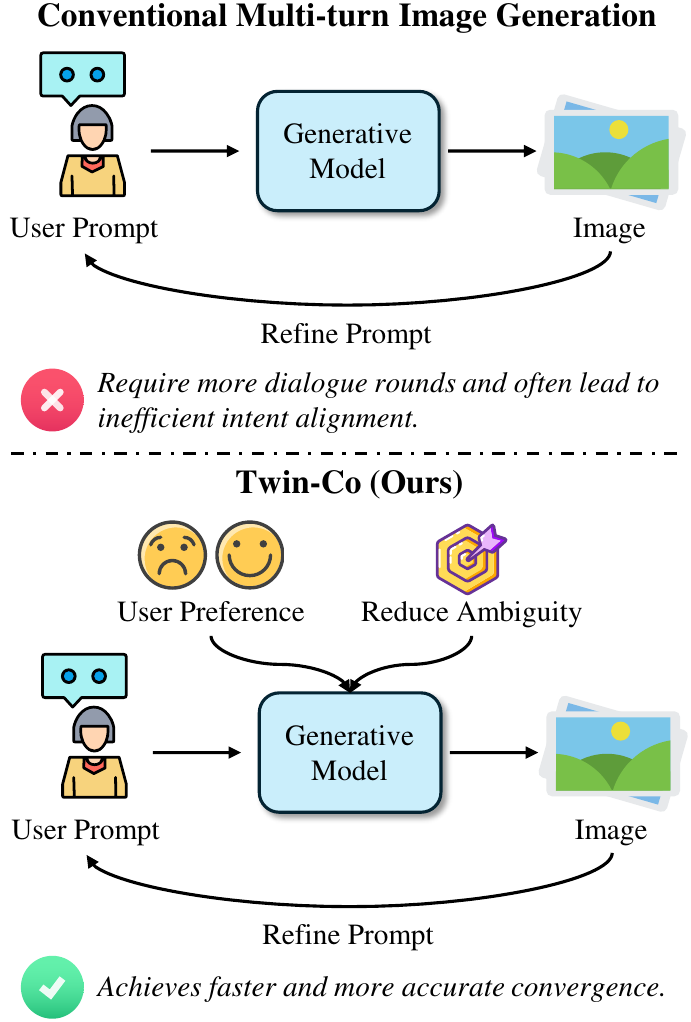} 
    \caption{Comparison between conventional multi-turn image generation and our proposed Twin-Co framework. Traditional approaches rely solely on iterative prompt refinement and often require more dialogue rounds to align with user intent. In contrast, Twin-Co integrates both explicit user feedback and implicit reflection mechanisms.}
    \label{fig:dialogue_example}
\Description{}\end{figure}

\section{Related Work}
\label{sec:related_work}

\subsection{Text-Driven Editing and Ambiguity Resolution}
Advances in text-to-image generation have emphasized improving alignment with user intent via dynamic feedback. For example, Hertz \emph{et al.}~\cite{hertz2022prompt} propose a framework that exploits cross-attention layers in diffusion models to perform prompt-guided image editing, achieving high-quality modifications. In parallel, ImageReward~\cite{xu2024imagereward} introduces reward models based on human preference feedback to fine-tune image-text alignment. Complementary approaches enhance this process by gathering both detailed and broad feedback signals~\cite{wu2023better, liang2023rich}. To further address prompt ambiguity, methods based on visual annotations~\cite{endo2023masked} and benchmarking protocols~\cite{lee2024holistic} have been developed, along with auto-regressive models~\cite{yu2022scaling} and masked transformer strategies~\cite{chang2023muse}. Notably, the TIED framework and TAB dataset~\cite{mehrabi2023resolving} leverage interactive feedback to clarify prompts, thereby boosting both precision and creativity in generated images.

\subsection{Human Feedback for Preference Alignment}
Incorporating human feedback to align generative outputs with user expectations has emerged as a key trend. The HIVE framework~\cite{10657924} fine-tunes a diffusion-based image editor using RLHF, where users rank multiple outputs to train a reward model that guides generation towards faithful adherence to instructions. Alternatively, a self-play fine-tuning strategy~\cite{yuan2024selfplay} allows the model to iteratively compare its outputs with prior versions, serving as its own teacher and achieving better alignment with human aesthetic and semantic preferences—while reducing the need for extensive human input. Furthermore, VisionPrefer~\cite{wu2024multimodal} utilizes multimodal language models to curate synthetic datasets of detailed image critiques that cover dimensions such as composition, style, and safety. By training reward models on these multi-dimensional signals, this approach substantially improves prompt alignment over traditional single-dimension feedback methods.

\subsection{Multi-turn Communication in Visual Generation}
Interactive dialogue systems have recently been introduced to iteratively refine image generation. For example, a proactive text-to-image agent~\cite{hahn2024proactiveagentsmultiturntexttoimage} tackles prompt underspecification by asking clarifying questions and maintaining an editable belief graph of user intent, thereby converging on the desired image through multi-turn interactions. Similarly, DiffChat~\cite{wang-etal-2024-diffchat} leverages a large language model to engage in multi-turn dialogue, refining prompts using reinforcement learning based on aesthetics, content integrity, and user preferences. In addition, PromptCharm~\cite{Wang_2024} and DialogGen~\cite{huang2024dialoggenmultimodalinteractivedialogue} also explore multi-turn communication by adopting alternative dialogue frameworks to enhance prompt clarity and semantic alignment. These studies demonstrate that integrating human-in-the-loop dialogue and feedback into the generation process yields more user-centered and responsive image generation systems.

In our earlier framework~\cite{he2025tdri}, we introduced a pose extraction in the initial generation phase. We treat this pose module as an optional enhancement. Experiments show that whether the pose constraint is enabled or disabled, the end‐to‐end performance on both automatic metrics and human preference remains unchanged. To simplify the pipeline and reduce runtime overhead for applications that do not demand strict pose control, we omit the pose extractor by default while retaining it as a plug‐in option for use cases requiring precise structural fidelity.

\begin{figure*}[htbp]
  \centering
  \includegraphics[width=\textwidth]{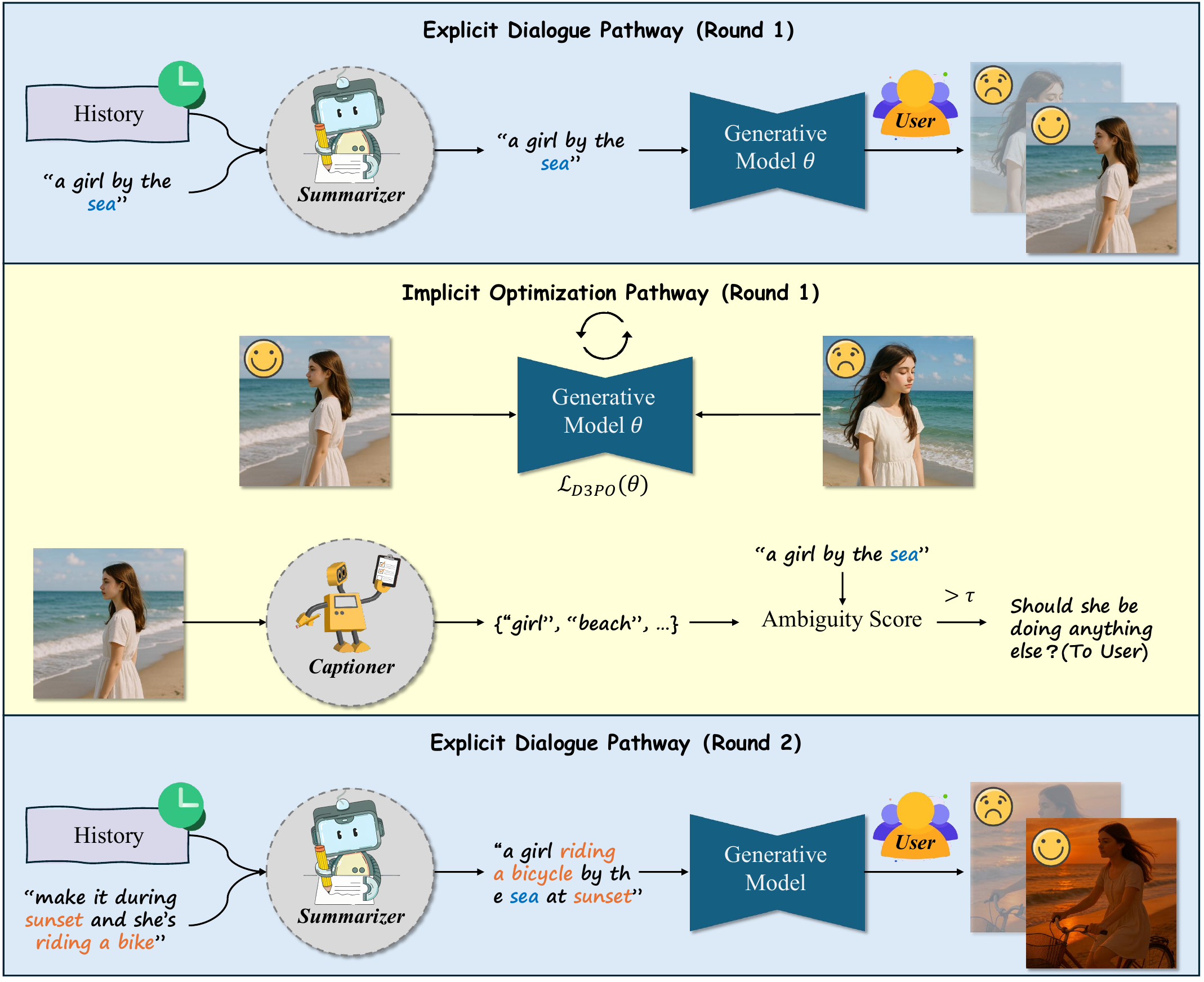} 
  \caption{Overview of the Twin-Co framework. 
\textbf{Top}: In the Explicit Dialogue Pathway (Round 1), the summarizer generates a prompt from dialogue history, which is used by the generative model to produce candidate images. The user provides feedback indicating preference. 
\textbf{Middle}: The Implicit Optimization Pathway leverages preference pairs for D3PO-based reward optimization and uses a captioner to extract semantic concepts from generated images. An ambiguity score is computed to determine whether clarification is needed. 
\textbf{Bottom}: In Round 2, the summarizer incorporates the user's clarification (e.g., ``riding a bike at sunset'') into the prompt, guiding the model toward more accurate image generation. This Twin-Co process is iteratively repeated over multi-turn dialogues to progressively align with user intent.}
  \label{Diagram}
\end{figure*}

\section{Method}
In this section, we introduce Twin-Co, a framework for reflective human-machine co-adaptation in multi-turn text-to-image generation. It is built to iteratively align generated images with user intent through both external dialogue and internal optimization.

\subsection{Problem Formulation}

Our goal is to progressively generate images that closely align with a user's intended visual concept through multi-turn dialogues. Formally, given an initial user-provided textual prompt \( w^{(1)} \), we aim to produce a final image \( I^* \) that precisely captures the user's intent, even if the initial description is incomplete or ambiguous. To systematically mitigate ambiguity in the user's intent, we propose a twin co-adaptive dialogue framework that simultaneously operates through two interconnected adaptive pathways:

\begin{itemize}
    \item \textbf{Explicit Dialogue Pathway:} This pathway focuses on explicit user interactions, refining prompt \(P^{(t)}\) based on direct user clarification at each dialogue turn.
    \item \textbf{Implicit Optimization Pathway:} Complementarily, this pathway utilizes internal model feedback—such as the semantic consistency between prompt \(P^{(t)}\) and image \(I^{(t)}\)—to optimize the generative process even without explicit user intervention.
\end{itemize}

By iteratively intertwining these two adaptive pathways, we progressively reduce the semantic gap between the generated images and the user's intended visual concepts, ensuring convergence toward the target image \(I^*\): $\lim_{t \to T} I^{(t)} \approx I^*$.

\subsection{Reflective Training under Twin Pathways}

Figure~\ref{Diagram} presents an overview of the Twin-Co training process. We begin by initializing our diffusion-based generative model through supervised fine-tuning on 2000 curated image-text pairs from the ImageReward dataset~\cite{xu2024imagereward}, where each pair \(\bigl(P, I^*\bigr)\) comprises a user-written prompt \(P\) and a high-quality reference image \(I^*\). These data serve as supervision for aligning textual cues with human-centric image semantics. Let \(\epsilon_\theta\) denote the denoising UNet parameterized by \(\theta\), and \(z_t\) be a noisy latent at timestep \(t\). Given a clean latent image \(z_0\), we apply a forward diffusion process 
\(
z_t \sim \mathcal{N}\bigl(\sqrt{\bar{\alpha}_t}\,z_0,\,(1 - \bar{\alpha}_t)\,\mathbf{I}\bigr)
\)
and train the model to predict the added noise \(\epsilon\) based on \(z_t, t,\) and \(P\). The standard diffusion loss is:
\begin{equation}
    \mathcal{L}_{\text{diff}}(\theta) 
    = \mathbb{E}_{z_0, \epsilon, t}\!
      \Bigl[\!\bigl\|\epsilon_\theta(z_t, t, P) - \epsilon\bigr\|^2\!\Bigr],
\end{equation}
where \(\epsilon \sim \mathcal{N}(0, \mathbf{I})\). This training objective drives the model to reconstruct clean latents conditioned on the prompt, yielding a robust initialization for subsequent dialogue-driven adaptation.

Next, we construct a multi-turn dialogue dataset to foster iterative user-model interactions. We select high-quality pairs \(\bigl(P, I^*\bigr)\) from the same dataset and simulate at least four dialogue rounds per pair by progressively refining the prompt \(P\). In each round, the model generates intermediate images and receives user-like feedback. Altogether, we accumulate over 2000 prompt variations, forming multi-turn dialogue traces that capture evolving user intent. Building upon these traces, we fine-tune the model via two complementary pathways: an \emph{Explicit Dialogue Pathway} driven by direct user inputs, and an \emph{Implicit Optimization Pathway} that leverages internal reflection on prompt-image consistency without explicit external supervision.

\noindent\textbf{Explicit Dialogue Pathway.} During training, at each dialogue turn \( t \), the model records the history:
\begin{equation}
    \mathcal{H}^{(t)} = \{(w^{(1)}, r^{(1)}), (w^{(2)}, r^{(2)}), \dots, (w^{(t-1)}, r^{(t-1)})\},
\end{equation}
where \( w^{(t)} \) denotes the user input at turn \( t \), and \( r^{(t)} \) denotes the system response at turn \( t \). 
We utilize a summarization module \( \mathcal{F}_P \) to produce a concise yet comprehensive prompt representation:
\begin{equation}
    P^{(t)} = \mathcal{F}_P\left(\mathcal{H}^{(t)}, w^{(t)}\right).
\end{equation}
Here, \(\mathcal{F}_P\) is implemented using GPT-4~\cite{openai2024gpt4technicalreport}. Specifically, the entire dialogue history \(\mathcal{H}^{(t)}\) together with the current user input \(w^{(t)}\) is provided as input to GPT-4, which generates a refined textual prompt \( P^{(t)} \). The updated prompt \(P^{(t)}\) is subsequently used to condition the generative model \(\mathcal{G}\) to synthesize a new image:
\begin{equation}
    I^{(t)} = \mathcal{G}(P^{(t)}).
\end{equation}
Through this iterative process, the Explicit Dialogue Pathway incrementally refines the internal prompt representation, ensuring that each generated image progressively aligns more closely with the user's evolving intent while maintaining overall contextual coherence.

\noindent\textbf{Implicit Optimization Pathway.} Parallel to explicit interactions, the \textit{Implicit Optimization Pathway} internally assesses and refines image generation quality through adaptive optimization without direct user intervention. Specifically, after generating each intermediate image \( I^{(t)} \), we employ a pretrained vision-language model (Qwen-VL (7B)~\cite{Qwen-VL}), denoted as \(\mathcal{E}(\cdot)\), to generate a comprehensive semantic caption set \( C^{(t)} \):
\begin{equation}
    C^{(t)} = \mathcal{E}\left(I^{(t)}\right) = \{C^{(t)}_1, C^{(t)}_2, \dots, C^{(t)}_N\},
\end{equation}
where each \(C^{(t)}_i\) captures a distinct semantic component of the generated image.

To quantify the alignment between the intended prompt representation \(P^{(t)}\) and the visual content described by \(C^{(t)}\), we define an ambiguity metric \(\delta^{(t)}\) based on the CLIP score~\cite{radford2021learningtransferablevisualmodels}:
\begin{equation}
    \delta^{(t)} = 1 - \frac{1}{N}\sum_{i=1}^{N}\text{CLIP}\left(P^{(t)}, C^{(t)}_i\right),
\end{equation}
where \(\text{CLIP}(P^{(t)}, C^{(t)}_i)\) denotes the cosine similarity computed between the feature embeddings of \(P^{(t)}\) and \(C^{(t)}_i\). When \(\delta^{(t)}\) exceeds a predefined threshold \(\tau\), an internal clarification strategy is triggered to generate a targeted clarification question:
\begin{equation}
    q^{(t+1)} = \mathcal{Q}\left(P^{(t)}, C^{(t)}, \delta^{(t)}\right),
\end{equation}
with \(\mathcal{Q}(\cdot)\) constructing questions aimed at resolving the most ambiguous aspects of the user's intent.

To further ensure text–image alignment, we apply the Attend-and-Excite~\cite{chefer2023attendandexciteattentionbasedsemanticguidance} loop immediately after obtaining \(I^{(t)}\).

Let \(I^{(t)}_1 = I^{(t)}\) and initialize the activation list \(\mathcal{T}_0 = \emptyset\). For each iteration \(n=1,2,\dots,N\), we compute  
\begin{equation}
\mathrm{Sim}_n \;=\;\mathrm{CLIP}\bigl(I^{(t)}_n,\,P^{(t)}\bigr).
\end{equation}
If \(\mathrm{Sim}_n \ge k\), we stop and output \(I^{(t)}_n\). Otherwise, we set the loss
\begin{equation}
\ell_n = 1 - \mathrm{Sim}_n
\end{equation}
and backpropagate \(\ell_n\) to the prompt to obtain the gradient  
\begin{equation}
\Delta P^{(t)}_n = \nabla_{P^{(t)}}\,\ell_n.
\end{equation}
We then pick the single token whose gradient magnitude is largest:
\begin{equation}
i^* = \arg\max_i \bigl|\Delta P^{(t)}_n[i]\bigr|
\end{equation}
and update the activation list
\begin{equation}
\mathcal{T}_n = \mathcal{T}_{n-1} \;\cup\;\{\,i^*\,\}.
\end{equation}
Using this cumulative set, we resample the image via Attend-and-Excite:
\begin{equation}
I^{(t)}_{n+1} = \mathrm{A\&E}\bigl(P^{(t)},\,\mathcal{T}_n\bigr).
\end{equation}
This loop repeats until \(\mathrm{Sim}_n\ge k\) or \(n=N\), ensuring that any tokens initially “neglected” by the diffusion model receive progressively stronger attention during sampling—without ever updating the model weights.

Furthermore, to autonomously optimize the generative process based on user preferences, we enhance our framework with a multi-step preference optimization mechanism called D3PO~\cite{yang2024usinghumanfeedbackfinetune}. Compared to traditional Direct Preference Optimization (DPO)~\cite{rafailov2024directpreferenceoptimizationlanguage}, which updates model parameters based solely on final outputs, D3PO treats the diffusion process as a multi-step Markov Decision Process (MDP). By incorporating user preferences at every denoising step, the model can more precisely adapt image generation toward user-favored outcomes. Formally, suppose that from multiple dialogue turns we gather preference pairs \(\{(I^{+}, I^{-})\}\). During the denoising process, let \((s^{+}, a^{+})\) denote the latent state and action for a preferred sample, and \((s^{-}, a^{-})\) for a non-preferred sample. We initialize a frozen reference model \(\pi_{\text{ref}}\) from \(\pi_\theta\), and then optimize \(\pi_\theta\) using the D3PO objective:
\begin{equation}
\label{eq:D3PO}
\mathcal{L}_{\text{D3PO}}(\theta) 
= 
- \mathbb{E}\!
\left[
  \log \rho
  \Bigl(
    \beta \log \frac{\pi_\theta(a^{+}\mid s^{+})}{\pi_{\text{ref}}(a^{+}\mid s^{+})}
    \;-\;
    \beta \log \frac{\pi_\theta(a^{-}\mid s^{-})}{\pi_{\text{ref}}(a^{-}\mid s^{-})}
  \Bigr)
\right],
\end{equation}
where \(\beta\) is a temperature parameter controlling divergence from \(\pi_{\text{ref}}\). This design enables the model to iteratively refine each denoising step in accordance with user-liked samples. In practice, we collect preference pairs by generating two candidates per round (as illustrated in Figure~\ref{Diagram}).

\subsection{User-Guided Inference}
During inference, as the generative model’s parameters have already been optimized via both explicit and implicit pathways during training. This yields a more lightweight inference process, suitable for real-world user interactions:

\begin{enumerate}
    \item \textbf{Dialogue Recording:} The system stores user inputs and past system responses in the dialogue history \(\mathcal{H}^{(t)}\). 
    \item \textbf{Prompt Summarization:} A summarizer \(\mathcal{F}_P\) encodes \(\mathcal{H}^{(t)}\) and the current user query \(w^{(t)}\), generating an updated prompt \(P^{(t)}\).
    \item \textbf{Image Generation:} The diffusion model \(\mathcal{G}\) produces the output image \(I^{(t)} = \mathcal{G}(P^{(t)})\), reflecting the most recent user directives.
\end{enumerate}

At this stage, no implicit optimization modules (such as D3PO) are invoked. By restricting inference to the explicit pathway, the system avoids additional computational overhead and responds quickly to user requests, making multi-turn dialogues more accessible for non-expert users.

\begin{table*}[!t]
\centering
\renewcommand{\arraystretch}{1.3} 
\caption{Comparisons of prompt-intent and image-intent alignment, supplemented by human evaluation. ``Augmentation'' refers to using LLMs to refine the initial prompt.}
\label{tab:baseline}
\resizebox{\textwidth}{!}{
\begin{tabular}{lccccc}
\toprule
\multirow{2}{*}{\textbf{Methods}} & \multicolumn{2}{c}{\textbf{Prompt-Intent Alignment}} & \multicolumn{2}{c}{\textbf{Image-Intent Alignment}} & \multirow{2}{*}{\textbf{Human Voting}} \\
\cmidrule(lr){2-3} \cmidrule(lr){4-5}
 & \textbf{T2I CLIPscore} & \textbf{T2I BLIPscore} & \textbf{I2I CLIPscore} & \textbf{I2I BLIPscore} & \\
\midrule
\multicolumn{6}{l}{\textbf{\emph{LLM-based Prompt Augmentation}}} \\
GPT-3.5 Augmentation                    & 0.154 & 0.146 & 0.623 & 0.634 & 5\% \\
GPT-4 Augmentation~\cite{openai2024gpt4technicalreport}                      & 0.162 & 0.151 & 0.647 & 0.638 & 6.2\% \\
LLaMA-2 Augmentation~\cite{touvron2023llama}                    & 0.116 & 0.133 & 0.591 & 0.570 & 4.1\% \\
Yi-34B Augmentation~\cite{ai2024yi}                     & 0.103 & 0.124 & 0.586 & 0.562 & 4.3\% \\
\midrule
\multicolumn{6}{l}{\textbf{\emph{No Interactive Refinement}}} \\
From Scratch Generation       & 0.100 & 0.110 & 0.570 & 0.560 & 3\% \\
Only Implicit Optimization & 0.220 & 0.231 & 0.712 & 0.693 & 12\% \\
\midrule
\multicolumn{6}{l}{\textbf{\emph{Interactive Refinement}}} \\
Explicit Dialogue Pathway Only           & 0.281 & 0.285 & 0.753 & 0.767 & 25.8\% \\
Explicit + ImageReward RL~\cite{xu2024imagereward}  & 0.297 & 0.284 & 0.786 & 0.776 & 26.5\% \\
\rowcolor{gray!20}
\textbf{Twin-Co (Ours)}                  & \textbf{0.338} & \textbf{0.336} & \textbf{0.812} & \textbf{0.833} & \textbf{33.6\%} \\
\bottomrule
\end{tabular}
}
\end{table*}

\section{Experiments}

We evaluate Twin-Co on the general text-to-image generation task using a diverse image–text dataset. This evaluation focuses on verifying the framework’s ability to understand and iteratively refine broadly varying user intents across multiple interaction rounds. Our experiments show that combining multi-turn dialogue with integrated internal optimization progressively refines generated images, yielding outputs that more accurately capture evolving user intent.

\subsection{Experimental Setup}
For our primary evaluation, we employ a curated subset of the ImageReward dataset~\cite{xu2024imagereward} containing high-quality image-text pairs. For each initial user prompt \(w^{(1)}\), we simulate realistic multi-turn dialogues consisting of at least four rounds, resulting in over 2000 refined prompt variations. In each round, our GPT-4-based summarizer aggregates the dialogue history into an updated prompt \(P^{(t)}\), which is then used to condition a diffusion-based generative model (Stable Diffusion v1.4~\cite{rombach2022highresolutionimagesynthesislatent} with DDIM~\cite{song2022denoisingdiffusionimplicitmodels} sampling) to produce an image \(I^{(t)}\). The generated images are further evaluated by the Qwen-VL model~\cite{Qwen-VL} to extract semantic descriptions. Alignment between the refined prompt, generated image, and a target reference image is quantified using CLIP\cite{radford2021learning} and BLIP~\cite{li2022blipbootstrappinglanguageimagepretraining} scores. All experiments were conducted on four Nvidia A6000 GPUs.

\begin{figure*}[t]
  \centering
  \includegraphics[width=0.95\textwidth]{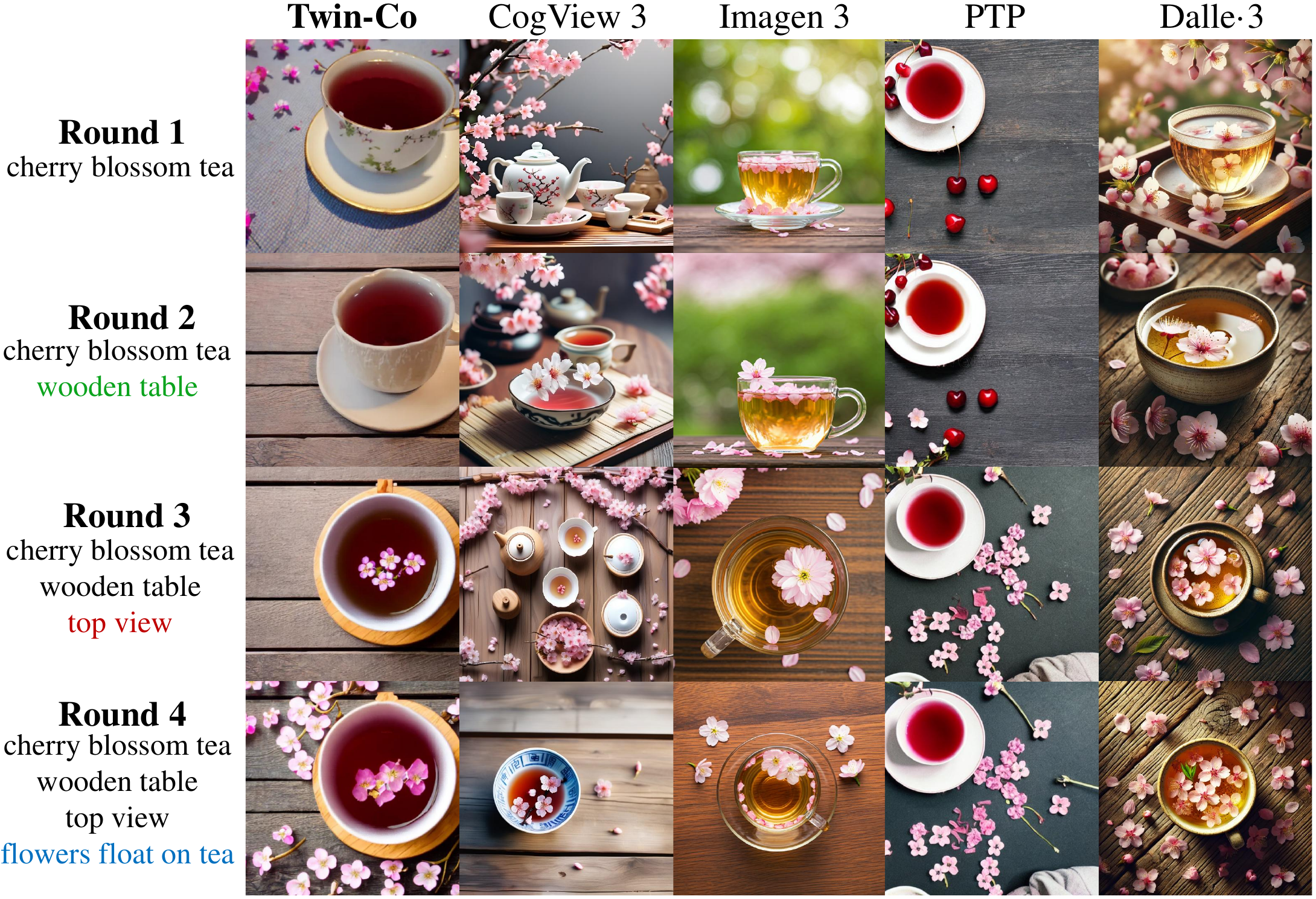} 
  \caption{Comparison of cherry blossom tea images generated across four dialogue rounds by various models.}
  \label{fig:visualization}
\end{figure*}

\subsection{Baseline Comparisons}

We compare our Twin-Co framework with several baselines, which we organize into three categories:

\textbf{(1) LLM-based Prompt Augmentation:} These methods employ large language models (LLMs) such as GPT-3.5, GPT-4~\cite{achiam2023gpt}, LLaMA-2~\cite{touvron2023llama} and Yi-34B~\cite{ai2024yi} to reformulate the initial user prompt prior to image generation. Although these approaches help reduce surface-level ambiguity, they do not incorporate iterative refinement or internal optimization, resulting in relatively modest performance (e.g., Yi-34B achieves a T2I CLIPscore of 0.103).

\textbf{(2) No Interactive Refinement:} This category includes methods that do not leverage multi-turn dialogue. In the From Scratch Generation (0-turn) baseline, the image is generated directly from the initial prompt without any subsequent refinement. The Only Implicit Optimization variant, on the other hand, relies solely on internal optimization mechanisms (A\&E combined with D3PO) without any explicit user feedback. These methods serve as a lower bound, with From Scratch Generation achieving a T2I CLIPscore of 0.100 and Only Implicit Optimization reaching 0.220.

\textbf{(3) Interactive Refinement:}  
These methods exploit multi-turn dialogue to iteratively update the prompt and guide image generation. For instance, the Explicit Dialogue Pathway Only baseline, which relies solely on direct user feedback over multiple rounds, achieves a T2I CLIPscore of 0.281. Further improvements are observed when reinforcement learning-based feedback is incorporated in the Explicit + ImageReward RL variant~\cite{xu2024imagereward}, which attains a T2I CLIPscore of 0.297. In contrast, our full Twin-Co framework—combining both explicit dialogue refinement and implicit internal optimization—achieves the best performance, with a T2I CLIPscore of 0.338, an I2I CLIPscore of 0.812, and a human voting preference of 33.6\%.

These results, summarized in Table~\ref{tab:baseline}, demonstrate that while external prompt augmentation and non-interactive methods provide limited performance gains, the unified dual-path adaptation embodied by Twin-Co significantly enhances both prompt-intent and image-intent alignment.

\subsection{Qualitative Analysis}

\textbf{Visual Comparison.}  
Figure~\ref{fig:visualization} illustrates a qualitative comparison of multi-turn image generation for the prompt "cherry blossom tea" across four dialogue rounds. As user intent becomes progressively refined—with added specifications like "wooden table," "top view," and "flowers float on tea"—our Twin-Co framework consistently produces visually coherent and semantically aligned images. In contrast, baseline models such as CogView 3~\cite{zheng2024cogview3finerfastertexttoimage}, Imagen 3~\cite{imagenteamgoogle2024imagen3}, PTP~\cite{hertz2022prompt}, and DALL·E 3~\cite{ramesh2022hierarchicaltextconditionalimagegeneration} struggle to incorporate fine-grained revisions. For instance, in later rounds, these models often fail to maintain consistent perspectives or to correctly render compositional cues like flower placement or camera angle. Imagen 3 and DALL·E 3 tend to generate aesthetically pleasing but semantically static images, while PTP and CogView 3 introduce distractive artifacts or ignore spatial refinements. In contrast, Twin-Co precisely integrates user feedback at each turn, capturing evolving visual semantics.

\subsection{User Study}  
To analyze how users interact with our system, we conducted a user study involving 20 participants, each completing 30 multi-turn sessions, yielding a total of approximately 600 dialogue traces. In each session, the user iteratively refined their prompt toward a target visual goal, with satisfaction recorded at each round.

\begin{figure}[htbp]
    \centering
    \includegraphics[width=\linewidth]{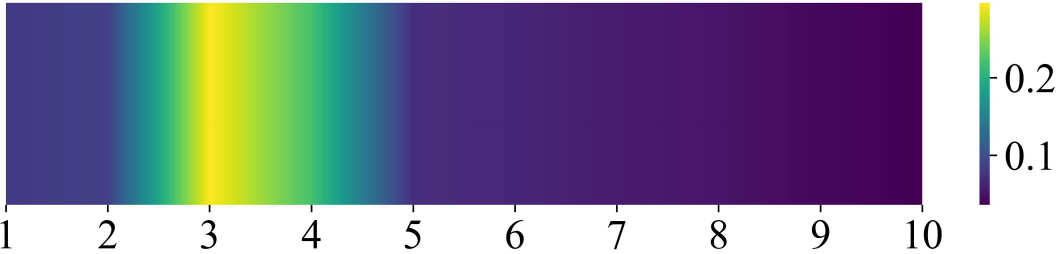}
    \caption{Heatmap showing user perception of intent capture across dialogue rounds. The intensity peaks around the third round.}
    \label{fig:heatmap_intent}
\Description{}\end{figure}

Figure~\ref{fig:heatmap_intent} presents a heatmap summarizing user perception of intent capture across dialogue rounds. The intensity peaks around the third round, indicating that users typically felt their intent was best understood at this stage. This aligns with the design of our ambiguity-aware clarification and preference-driven optimization mechanisms, which begin to show noticeable effects early in the interaction process.

\begin{figure}[t]
    \centering
    \resizebox{\columnwidth}{!}{
    \includegraphics[width=\linewidth]{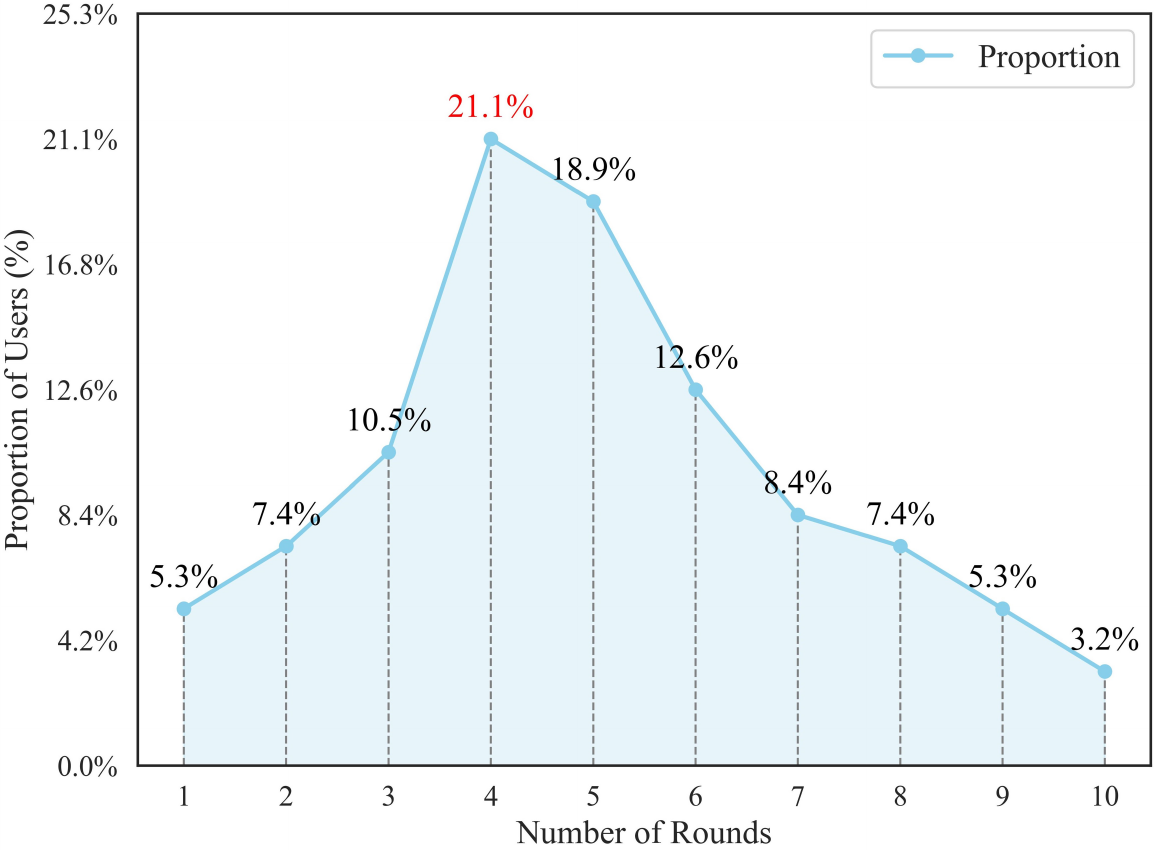} }
    \caption{User interaction distribution across dialogue rounds. The peak at 4 rounds (21.1\%) indicates that most users achieve satisfactory results within 4 interactions.}
    \label{fig:rounds_proportion}
\Description{}\end{figure}

Figure~\ref{fig:rounds_proportion} illustrates the distribution of user interactions needed to achieve satisfactory results. The curve peaks at 4 rounds (21.1\%), showing that most users completed their refinement within a small number of interactions. These results suggest that Twin-Co facilitates efficient co-adaptation between human intent and machine generation, enabling fast convergence toward the desired outcome.

\subsection{Ablation Studies}  
\textbf{Two-Pathway Ablation.}  
Table~\ref{tab:baseline} includes ablations isolating each adaptation pathway. The “Only Implicit Optimization” setting excludes user interaction, while “Explicit Dialogue Pathway Only” omits internal optimization. Twin-Co, combining both, achieves the best results confirming the complementary strengths of explicit and implicit adaptation.

\noindent\textbf{Attend-and-Excite Threshold Analysis.}  
Table~\ref{tab:tool2_thresholds} examines the impact of different threshold values on the usage frequency and effectiveness of the Attend-and-Excite (A\&E) module. The threshold controls how aggressively the system reactivates under-attended tokens in the prompt. As the threshold lowers from 0.80 to 0.66, the usage frequency of A\&E increases—from 0\% to 95.8\%—indicating that more prompts are identified as semantically under-aligned. At the same time, the T2I similarity improvement (measured by CLIP score gain) first increases, peaking at a threshold of 0.68 with a 2.67\% gain, and then slightly drops when the threshold is too low (1.3\% at 0.66). This pattern reveals a trade-off: while more aggressive reactivation improves recall of prompt details, it may also introduce noise if applied indiscriminately. The results confirm that the Attend-and-Excite module is most effective when selectively activated, and underscore the importance of adaptive internal mechanisms in refining image-text alignment. It supports our design choice to embed A\&E within the implicit optimization pathway of Twin-Co.

\begin{table*}[t]
    \centering
    \caption{Attend-and-Excite usage frequency and T2I similarity at different thresholds.}
    \label{tab:tool2_thresholds}
    \small
    \begin{tabular}{@{} l *{6}{c} @{}}
        \toprule
        \textbf{Attend-and-Excite Threshold} & \textbf{0.80} & \textbf{0.75} & \textbf{0.73} & \textbf{0.70} & \textbf{0.68} & \textbf{0.66} \\
        \midrule
        \textbf{Frequency of Usage} & 0 & 8.7~\% & 31.3~\% & 51.6~\% & 72.5~\% & 95.8~\% \\
        \textbf{T2I Similarity Improvement} & 0 & 0.23~\% & 1.87~\% & 2.36~\% & 2.67~\% & 1.3~\% \\
        \bottomrule
    \end{tabular}
\end{table*}

\begin{table*}[!t]
\centering
\caption{Comparison across ablation experiments. (Best results in \textbf{bold})}
\label{tab:merged_comparison}
\resizebox{\textwidth}{!}{
\begin{tabular}{llcccccccc}
\toprule
 & & \textbf{Consistency} & \textbf{Gen. Success} & \textbf{T2ICLIP} & \textbf{User Sat.} & \textbf{Human} & \textbf{Conv.} & \textbf{Time} \\
\textbf{Category} & \textbf{Method/Setting} & \textbf{Score} & \textbf{Rate (\%)} & \textbf{Score} & \textbf{(\%)} & \textbf{Vote (\%)} & \textbf{Iters} & \textbf{(min)} \\
\midrule
\multirow{2}{*}{\textbf{Image Editing vs. From Scratch}} 
   & From Scratch          & 0.75 & --   & --   & 78 & --   & -- & 12 \\
   & Image Editing         & \textbf{0.88} & --   & --   & \textbf{90} & --   & -- & \textbf{9}  \\
\midrule
\multirow{2}{*}{\textbf{Simple vs. Complex Prompts}} 
   & Simple Prompts        & --   & \textbf{92}   & \textbf{0.27} & -- & \textbf{87}   & -- & -- \\
   & Complex Prompts       & --   & 65   & 0.16 & -- & 62   & -- & -- \\
\midrule
\multirow{2}{*}{\textbf{Generalized vs. Sample-Specific D3PO}}     
   & Generalized Model     & --   & --   & 0.22 & 83 & --   & \textbf{5}  & -- \\
   & Sample-Specific Model & --   & --   & \textbf{0.26} & \textbf{90} & --   & 8  & -- \\
\midrule
\multirow{4}{*}{\textbf{Effect of Interaction Turns}}               
   & 2  & --   & --   & 0.18 & 70 & --   & -- & \textbf{6}  \\
   & 4  & --   & --   & 0.24 & 85 & --   & -- & 9  \\
   & 6  & --   & --   & 0.29 & 87 & --   & -- & 11 \\
   & 8  & --   & --   & \textbf{0.34} & \textbf{88} & --   & -- & 12 \\
\bottomrule
\end{tabular}}
\end{table*}

\noindent\textbf{Image Editing vs. From-Scratch Generation.}  
As shown in Table~\ref{tab:merged_comparison} under the “Image Editing vs. From Scratch” category, the “From Scratch Generation (0-turn)” baseline achieves a consistency score of 0.75 with a user satisfaction of 78\% and an inference time of 12 minutes. In contrast, the “Image Editing” method yields a higher consistency of 0.88, improved user satisfaction of 90\%, and a reduced generation time of 9 minutes. These results clearly indicate that iteratively refining an existing image leads to more robust alignment with user intent and enhanced computational efficiency.

\noindent\textbf{Simple vs. Complex Prompts.}  
In the “Simple vs. Complex Prompts” category of Table~\ref{tab:merged_comparison}, simple prompts achieve a generation success rate of 92\% and a CLIP score of 0.27, with 87\% of human votes favoring the results. Conversely, complex prompts obtain only a 65\% success rate and a lower CLIP score of 0.16, with 62\% in human voting. This substantial performance gap validates our decision to exclude overly complex prompts, as their inherent ambiguity undermines generation quality.

\noindent\textbf{Generalized vs. Sample-Specific D3PO.}  
According to the “Generalized vs. Sample-Specific D3PO” results in Table~\ref{tab:merged_comparison}, the Sample-Specific Model achieves a higher CLIP score (0.26) and user satisfaction (90\%) compared to the Generalized Model (0.22 and 83\%, respectively), albeit with a longer convergence (8 iterations versus 5). This indicates that while fine-tuning for individual dialogue contexts can deliver better performance, it does so with increased computational overhead.

\noindent\textbf{Effect of Interaction Turns.}  
The “Effect of Interaction Turns” section of Table~\ref{tab:merged_comparison} shows that increasing dialogue rounds leads to steady improvements in image quality and user satisfaction. Specifically, as the number of rounds increases from 2 to 8, the CLIP score rises from 0.18 to 0.34 and user satisfaction improves from 70\% to 88\%, although the incremental gain diminishes beyond 6 rounds and the time cost increases from 6 to 12 minutes.

Collectively, these ablation studies, as summarized in Table~\ref{tab:merged_comparison}, demonstrate that while basic prompt augmentation and non-interactive strategies yield limited improvements, the integrated dual-path adaptation in Twin-Co—combining explicit multi-turn dialogue with fine-grained implicit internal optimization—significantly enhances both prompt-intent and image-intent alignment.

\section{Conclusion}
We presented Twin-Co, a dual-path co-adaptive framework that combines explicit multi-turn dialogue with implicit internal optimization to enhance text-to-image generation. By iteratively refining prompts through user interaction and integrating techniques such as ambiguity detection, Attend-and-Excite, and preference-based optimization, Twin-Co effectively aligns generated outputs with user intent. Our extensive experiments across general and domain-specific settings demonstrate the framework's superiority over existing baselines in both quantitative metrics and human evaluations. Moving forward, we plan to explore broader deployment scenarios to further advance interactive visual content generation.

\bibliographystyle{ACM-Reference-Format}
\balance
\bibliography{sample-base}

\end{document}